# A Three-dimensional Convolutional-Recurrent Network for Convective Storm Nowcasting

Wei Zhang[1], Wei Li[1], Lei Han[1]

[1]College of Information Science and Engineering, Ocean University of China, Qingdao, Shandong, China

Abstract

Very short-term convective storm forecasting, termed nowcasting, has long been an important issue and has attracted substantial interest. Existing nowcasting methods rely principally on radar images and are limited in terms of nowcasting storm initiation and growth. Real-time re-analysis of meteorological data supplied by numerical models provides valuable information about three-dimensional (3D), atmospheric, boundary layer thermal dynamics, such as temperature and wind. To mine such data, we here develop a convolution-recurrent, hybrid deep-learning method with the following characteristics: (1) the use of cell-based oversampling to increase the number of training samples; this mitigates the class imbalance issue; (2) the use of both raw 3D radar data and 3D meteorological data re-analyzed via multi-source 3D convolution without any need for handcraft feature engineering; and (3) the stacking of convolutional neural networks on a long short-term memory encoder/decoder that learns the spatiotemporal patterns of convective processes. Experimental results demonstrated that our method performs better than other extrapolation methods. Qualitative analysis yielded encouraging nowcasting results.

*Keywords— Convective storm nowcasting, Convolutional neural Networks，Long short-term memory*

## I. INTRODUCTION

Very short-term, strong, convective weather nowcasting seeks to predict convective weather in the near future (e.g., 0–2 h) [1-3]. Strong convective weather is a common type of severe weather, being sudden, of short duration, and destructive. It is difficult to take effective protective measures over a short period of time; such weather poses severe risks to life and property [4]. Convective storm nowcasting has attracted substantial interest [5-8]. Accurate prediction has many real-world applications in agriculture, as well as outdoor activities, aviation, and power transmission [9].

Existing methods include extrapolations, conceptual model-based methods, and numerical weather predictions (NWP) [10]. Extrapolation employs principally reflectivity echoes detected by weather radar to predict storm speeds and directions. A classic example is the Thunderstorm Identification, Tracking and Nowcasting algorithm (TITAN), which has been widely used worldwide to track single convective storms [11]. However, when the radar echoes of strong convection processes change rapidly, forecasting accuracy plummets. The Auto-Nowcast system combines and analyzes a variety of meteorological data when engaging in conceptual forecasting [12]. Numerical model forecasting solves physical equations predicting atmospheric motions and weather phenomena. Although numerical predictions are widely used for mid- and long-term weather forecasting, such methods remain inapplicable over the short term because of the spin-up problem and an inadequate scientific understanding of convective processes.

Although existing nowcasting methods rely principally on radar observations, real-time re-analysis of meteorological data can provide valuable information about atmospheric, boundary layer thermal dynamics, which can be used for nowcasting. Here, we use data provided by a state-of-the-art numerical weather re-analysis system: the Variational Doppler Radar Analysis System (VDRAS) of the National Center for Atmospheric Research (NCAR) of the United States. Recently, the use of machine- or deep-learning methods to extract useful information from large amounts of data has attracted much attention [13-20]. In [17], a traditional machine-learning method was used for re-analysis of meteorological data to facilitate nowcasting. With interesting results: convolutional neural networks (CNNs) efficiently and automatically extract spatial features in the absence of handcraft engineering [21-23], the authors of [22] developed a dynamic CNN using consecutive past radar images to produce future radar images. Inspired by this work, we here use CNNs to extract spatial features from multi-source 3D input data.

## II. RELATED WORK

Recently, recurrent NN models have provided new insights into the use of spatiotemporal data for predictions, based on modeling of historical observations. In particular, the authors of [24] viewed precipitation nowcasting problems as spatiotemporal sequence-forecasting issues; past radar maps were sequenced as inputs that output a sequence of future maps. Convolutional operations were used to capture spatiotemporal correlations between the input-to-state and state-to-state transitions of long short-term memory (LSTM) neurons. The authors of [25] went beyond conventional LSTMs when developing the Trajectory GRU (TrajGRU) model, which learns the location-varying structures of recurrent connections. The authors of [26] developed a PredRNN architecture, the core of which was a spatiotemporal LSTM (ST-LSTM) unit that simultaneously extracted and memorized spatial and temporal representations.

LSTMs can be used to process and predict events with short and long time-sequence dependencies [27-30]; CNNs usefully extract spatial information. Thus, we developed a 3D convolution-recurrent, hybrid deep-learning method that we term CNN_LSTM to exploit the strengths of the two methods. Our hybrid model features a CNN extracting spatial features from input data at three adjacent moments and an LSTM encoder/decoder predicting convective storms based on the spatial features of each time step. The input data include both raw 3D radar data and re-analyzed meteorological data (which contain more spatiotemporal information); the combination improves predictive accuracy.

Our contributions may be summarized as follows:

1) **Cell-based oversampling.** Given the low frequency and relatively short duration of convective weather, a typical class-imbalance learning problem is in play. We use cell-based oversampling to increase the proportion of positive samples in the training set. (section III C)

2) **Multi-source 3D convolution.** Real-time re-analysis of meteorological data yields valuable information on atmospheric boundary layer, thermal dynamic information; we use multi-source 3D convolution to extract information from both sets of data without any need for handcrafted feature engineering. (section IV A)

3) **Stack CNNs on the LSTM encoder/decoder.** The CNNs extract spatial features from input data at three adjacent moments, and then feed the sequenced features into the LSTM encoder/decoder. The output from the last state predicts whether a convective storm will occur within the next 30 min at a specific location. (section IV B)

Unlike existing models, we treat strong precipitation nowcasting as a binary classification problem: will the radar echo exceed 35 dBZ in a specific location 30 min from now? Radar reflectivity data serve as the ground truths. The sample labels are: if the radar value in 30 min exceeds 35 dBZ, 1; otherwise, 0.

## III. DATA PREPROCESSING

The radar reflectivity data (R data) are from the KFTG WSR-88D radar of Denver (CO, USA). The re-analysis real-time meteorological data that we re-analyzed are those of the VDRAS. As air parcel buoyancy, vertical motion, and time trends play essential roles in storm development, we use the VDRAS pt, dpt, w, and dw variables as inputs, where pt is the perturbation temperature (proportional to buoyancy), w is the vertical velocity, and dpt and dw are time trends. All data fed to the CNN_LSTM model are raw 3D data downloaded from the link of [17] and include information on seven historical heavy rainfall events (on 8–9

August 2008, 28–29 July 2010, 9–10 August 2010, 13–14 July 2011, 14–15 July 2011, 6–7 June 2012, and 7–8 July 2012). We used the method of [17] (SBOW) for comparison; this is a support vector machine-based model. The study domain is the Denver area (Fig. 1): the total area is 280 x 230 km².

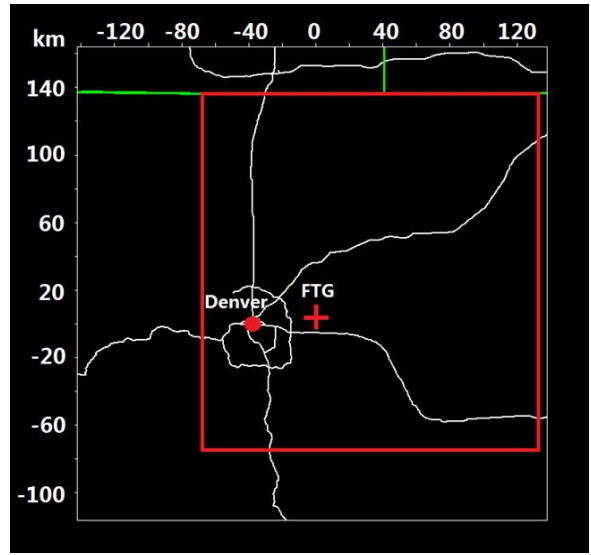

Fig. 1. The study domain (red rectangle) used in this study The green lines indicate the state borders and the white lines are the major highways.

## A. Data difference

Severe convective weather develops continuously. The present state of the atmosphere is affected by its previous state. Therefore, time trends must be considered. Temporal changes in each variable are required; we used a time interval of 15 min. For example, R information is obtained using the following formula:

$$dR = R_t - R_{t-15\min} \tag{1}$$

## B. Cell-based sampling

Unlike in other methods, we nowcast for small cells, not single pixels. We nowcast whether a storm will occur in a specific location/cell; such an approach is favored by forecasters. As weather phenomena are continuous in space (each cell is affected by all neighboring cells), we use 18- × 18-km windows centered on cells to consider spatial effects (Fig. 2). For each cell, six variables are input; each is a three-dimensional field (18×18×20 km³) with 20 2D slices. The projection of each variable is an 18×18 km window because each variable has 20 vertical layers. Therefore, for each cell at any moment, the sample dimensions are 6×18×18×20, as shown as Fig. 3.

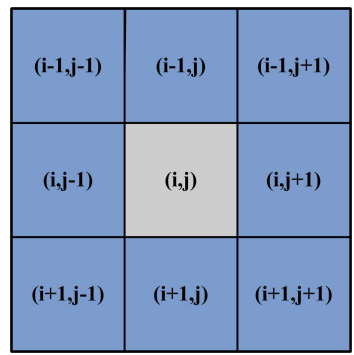

Fig. 2. Each data sample is collected based on 9 cells. Each cell is 6×6 km. The window (18×18km) also includes other 8 cells surround. The gray cell centered is what we need to nowcast.

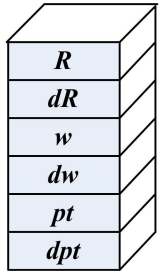

Fig. 3. The 3D data sample.

## C. Cell-based oversampling

Given the low frequency and relatively short duration of convective weather, we mitigated the class-imbalance learning issue. We use cell-moving oversampling to increase the number of positive samples in the training set (Fig. 4). If a cell is positive (gray in Fig. 4), the window moves by K (K = 1 or 2) pixels. If a new cell (red) in the new window is also positive, we add it to the training set. Via such oversampling, the ratio of positive samples was increased from the original 4% to 29.65%. However, in real scenarios, the test set maintains its original proportion; we do not oversample.

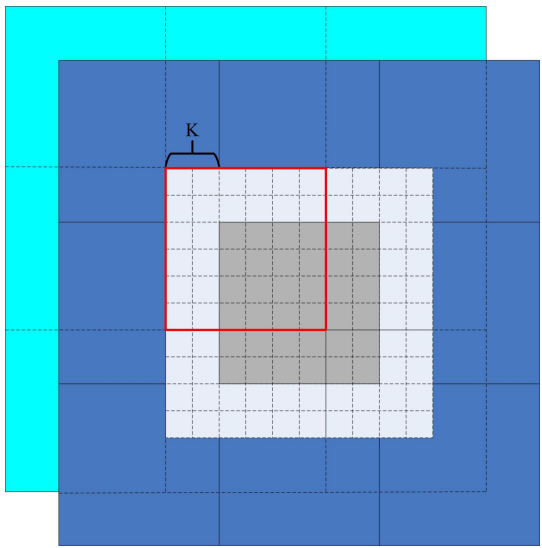

Fig. 4. The cell- moving oversampling strategy

## D. Data normalization

As VDRAS data are often in different units, we normalize them (0, 1) as follows. Seven instances of heavy convective weather (8–9 August 2008, 28–29 July 2010, 9–10 August 2010, 13–14 July 2011, 14–15 July 2011, 6–7 June 2012, and 7–8 July 2012) in the Denver area were used to train and test the model.

$$x^* = (\frac{x - x_{min}}{x_{max} - x_{min}}) \times 2 - 1 \qquad (2)$$

## IV. THE MODEL

### A. Multi-source 3D convolution

As described in section II, a data sample X features six collections of raw 3D raw data (Fig. 3). Each feature $X_i$ is a three-dimensional field (18×18×20) with 20 2D slices. The projection of $X_i$ is an 18×18-km window.

$$x = (x_1, x_2, x_3, x_4, x_5, x_6) = (R, dR, w, dw, pt, dpt) \qquad (3)$$

The six coupled physical variables play different roles but act together to control convective weather. It is thus essential to convolve the different variables; we used multisource 3D convolution, as described below. The first convolutional layer has 80 kernels. We establish the following equation:

$$X_k^1 = \text{ReLU}(\sum_{i,j} W_{ijk}^1 * X_{ij} + b_k^1), k = 1,\ldots,80 \qquad (4)$$

where i is the index of the variable, j is the index of a slice within a variable, k is the index of a feature map, $b_k^1$ is the linear bias, $X_k^1$ is the resulting feature map, $W_{ijk}^1$ is the weight matrix, and $X_{ij}$ is the input 2D slice of layer j within a variable. Examples of $X_{ij}$ and $W_{ijk}^1$ are shown on the left of Fig. 6. In general, multi-source variables can be 3D-convoluted in two different ways. Altitude information for each variable is important in terms of convective weather, so for each altitude, we convolved 2D slices onto different variables to generate a layer-feature map; we thus obtained 20 such maps that we then convolved to provide an overall feature map. This is one way to perform multi-source 3D convolution. However, because some 3D variables affect convective weather as a whole (and thus cannot be divided into 2D slices convolved by other variables), we convolved 2D slices onto different layers of the same variable to generate variable-feature maps, and then convolved all such maps onto an overall feature map. Note that equation 3 includes both of the above approaches, and that both can be automatically applied.

## B. Network architecture

We term our hybrid model CNN_LSTM. First, we use a CNN featuring multi-source 3D convolution to construct a 50-dimensional spatial vector. We then input these vectors for three adjacent time-steps into the LSTM model; the model thus learns temporal information. The stacked CNN_LSTM is then trained jointly (as a whole model), as shown in Fig. 5. Fig. 6 shows the CNN structure. Data inputs at each time are transformed into 50-dimensional vectors via four convolutions, followed by a single pooling and two full connection operations.

● **CNN layer**: The input data are convolved using several convolutional kernels. The kernel size is usually 5×5 in the first convolution layer and 3×3 in the later three convolution layers; rectangles provide vital local features. To accelerate both training and convergence, we normalize the data after convolution. Batch normalization follows, and the max-pooling layer reduces the sizes of all extracted feature maps. This in turn reduces the model parameters. We feed the data thus derived into the two fully connected layers to obtain a robust vector. The final fully connected layer has dimensions of 50×1, where 50 is the feature dimension. The entire process is shown in Fig. 6. After all operations are concluded for the three time-steps, we feed all three feature vectors into a concatenation layer to obtain a robust feature vector of dimensions 150×1.

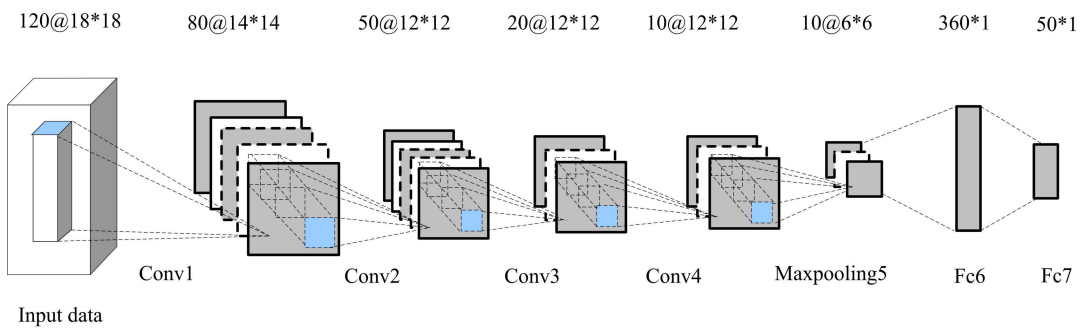

Fig. 5. The multi-source 3D convolutional CNN, the 50-dimenssional vector in layer Fc7 is the input of one time-step in LSTM.

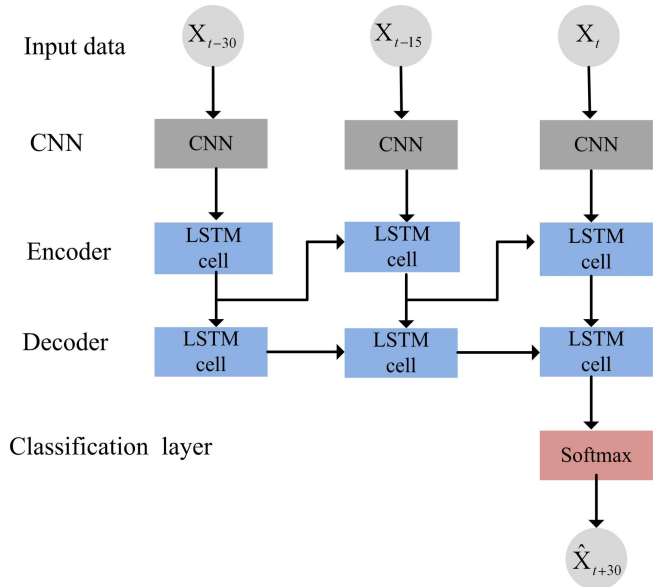

Fig. 6. The CNN_LSTM model

- **LSTM layer:** Prior to feeding the LSTM model, we change the compressed features of every moment into one-dimensional vectors with dimensions 150×1; each unit in the LSTM layer corresponds to a feature vector at a particular moment; the LSTM learns time information.

- **Classification layer:** We use a softmax layer as the final classifier.

## V. Experiments

### A. Evaluation criteria

The evaluation method used is based on the confusion matrix of Table I. A 35-dBZ radar threshold is used to convert nowcasting problems into a binary classification issue [5]. When nowcasting, 30-min intervals are often used to verify performance [11]; we also employ that interval.

TABLE I. Confusion Matrix

| **Confusion Matrix** | | |
|---|---|---|
| | Predicted Positive | Predicted Negative |
| Real Positive | True Positive (TP) | False Negative(FN) |
| Real Negative | False Positive(FP) | True Negative(TN) |

We calculate the probability of detection (POD), the false alarm ratio (FAR), and the critical success index (CSI); these metrics are similar to 'precision' and 'recall' (skill scores used in machine-learning), but we use three 'new' metrics compared to those two 'old' metrics. Previous studies have shown that when assessing low-frequency events such as convective weather, CSI POI and FAR are the best choice [17, 24-26, 31]. The definitions of POD, FAR, and CSI are:

$$POD = \frac{TP}{TP + FN} \quad (5)$$

$$FAR = \frac{FP}{TP + FP} \quad (6)$$

$$CSI = \frac{TP}{TP+FN+FP} \quad (7)$$

## B. Results and Analysis

As mentioned, we studied seven convective events, of which the first five were used for training and validation and the last two for testing (6 June 2012 and 7 July 2012; total sample no. about 60,450). We compared the CNN_LSTM model to several methods, not only the traditional and deep-learning methods:

- TITAN: The Thunderstorm Identification, Tracking and Nowcasting algorithm [11]. Today, TITAN nowcasts are cell-based.

- SBOW: The support vector machine method in [17].

- TrajGRU: A state-of-art model that actively learns location-variant structures required for precipitation nowcasting [25]. For convenience, we convert the TrajGRU results from images to cells to allow algorithm evaluation under the same conditions.

**Quantitative analysis.** Table II shows that the CNN_LSTM model performs better than the other methods in terms of attaining the highest CSI value. The TrajGRU model was trained using radar images, whereas the CNN-LSTM model used cell-based data for training. Further, a radar image can be divided into 1209 cells; the latter training set is thus 1209-fold larger than the former. Also, we oversampled the cell-based training set to balance the proportions of positive and negative samples. Using cell sampling, the CNN_LSTM can trace a storm trajectory accurately even when the dataset is small. A receiver operating characteristics (ROC) curve can be plotted and the area under the curve (AUC) value can be calculated based on the probability that a predicted result will eventuate. The TrajGRU calculates regression values lacking associated probabilities; TITAN also lacks probabilities. Thus, we could not plot ROCs for TITAN and TrajGRU. We implemented the SBOW method of [17]. The ROC curves of CNN_LSTM (Fig. 7) lie near the top left. The AUC is approximately 0.9366, much higher than the 0.7299 of SBOW; our model is better.

TABLE II. COMPARISION OF Four METHODS

| Model | CSI | POD | FAR |
|---|---|---|---|
| TITAN | 0.38 | 0.53 | 0.41 |
| SBOW | 0.39 | 0.61 | 0.47 |
| TrajGRU[1] | 0.41 | 0.50 | 0.33 |
| **CNN_LSTM** | **0.44** | **0.67** | **0.43** |

Fig. 8 presents the consecutive 30-min predicted results of CNN_LSTM. Over time, storms grow and the model performs better. Obviously, stronger storms are easier to predict. This is why the CSI is higher in the middle period. In each of the above cases, the storm becomes stronger as the POD rises, and the FAR becomes increasingly lower over time. In summary, the CSI rises with time, indicating that the CNN_LSTM improves predictive capacity.

---

[1] https://github.com/sxjscience/HKO-7

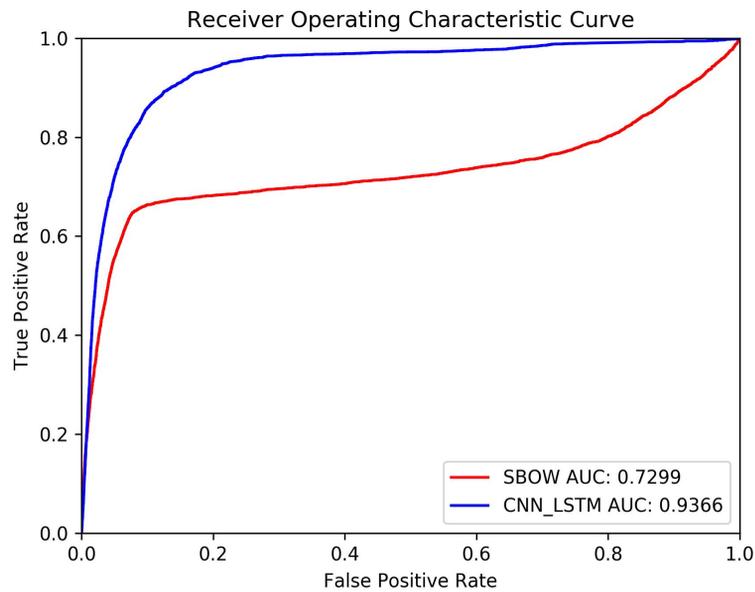

Fig. 7. The ROC curve

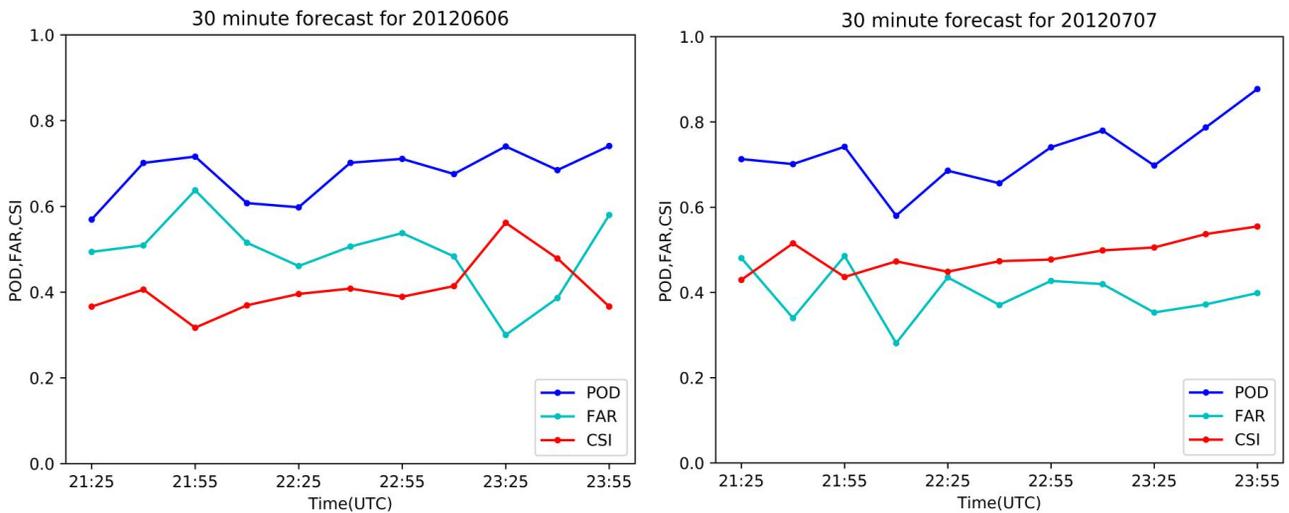

Fig. 8. Nowcasting statistics for 6 June and 7 July 2017

**Qualitative Analysis.** Here, we present the nowcasts afforded by different methods.

Fig. 9 presents four examples of convective storm nowcasts (one in each row). The left column contains the TrajGRU nowcasts, and the middle column the CNN_LSTM and TITAN nowcasts. The black cells represent correct predictions, the red cells represent false alarms, and the white cells represent missing forecasts. The cyan polygons are the TITAN nowcasts. The right column contains the ground truths. All Fig.s are superimposed on radar reflectivities. When comparing Fig. 9(a) and (b) at 23:55 UTC to the ground truth of Fig. 9(c), it is evident that the CNN_LSTM nowcasts capture storm advection better than do TrajGRU and TITAN. The results at 20:55, 21:10, and 21:25 UTC show that the CNN_LSTM nowcasts capture storm movement well. Both TrajGRU and TITAN also yield good forecasts, but both tend to underforecast, thus forecasting for smaller areas and exhibiting lower PODs, reflecting more misses (white cells).

As in Fig. 9, Fig. 10 presents three examples (in three rows) of line storm nowcasts. The left column contains the nowcasts of TrajGRU and the middle column represents the nowcasts of CNN_LSTM and TITAN. The black cells represent correctly predicted nowcasts, the red cells represent false alarms, and the white cells represent missing nowcasts. The cyan polygons are the TITAN nowcasts. The rightmost column is the ground truths. All Fig.s are superimposed on radar reflectivities. When comparing Fig. 10(a) and (b) at 00:10 UTC to the ground truths of Fig. 10(c), it is apparent that TrajGRU forecasts feature more areas lacking forecasts (white arrows) than do CNN_LSTM nowcasts. The last two rows [Fig.10(d–f) and Fig.10(g–i)] show that the CNN_LSTM captures line storm trends well; the data are in good agreement with the ground truths. TrajGRU and TITAN forecasts are more cautious and tend to underpredict storm areas.

Fig. 11 presents examples of storm initiation and rapid growth; the CNN_LSTM correctly predicts these events. As above, the black cells represent those predicted correctly, the red cells represent false alarms, and the white cells represent areas lacking forecasts. All Fig.s are superimposed on radar reflectivities. Fig. 11(a) (for 19:40 UTC) reveals no convective storm; Fig. 11(b) (at 19:55 UTC) shows the moment when a convective storm develops (i.e., storm initiation); the CNN_LSTM predicts this well. Fig. 11(c–h) (at 20:10, 20:25, 20:40 and 20:55, 21:10, 21:25 UTC, respectively) shows the nowcasts for storm growth over an increasing area; the CNN_LSTM captures this well. From beginning to end, the CNN_LSTM nowcasts are accurate, with few misses and false alarms.

## VI. CONCLUSION

We develop a hybrid convective weather model called CNN_LSTM to nowcast strong convective weather. First, we use a CNN to extract spatial features from input data derived at three adjacent time-steps. We then feed extracted spatial features into a temporal LSTM predicting strong convective storms in the next 30 min. The raw 3D input data include both radar information and re-analyzed meteorological data output by a numerical model; this allows nowcasting to employ more spatiotemporal information. We use cell-based oversampling to mitigate the class imbalance issue; this improves predictive accuracy. Experiments show that our model outperforms other models, well capturing the spatiotemporal trends of convective weather.

## ACKNOWLEDGMENT

This work was supported jointly by National Key R&D Program of China (grant 2018YFC1507504-6) and the National Natural Science Foundation of China (grant 41875049).

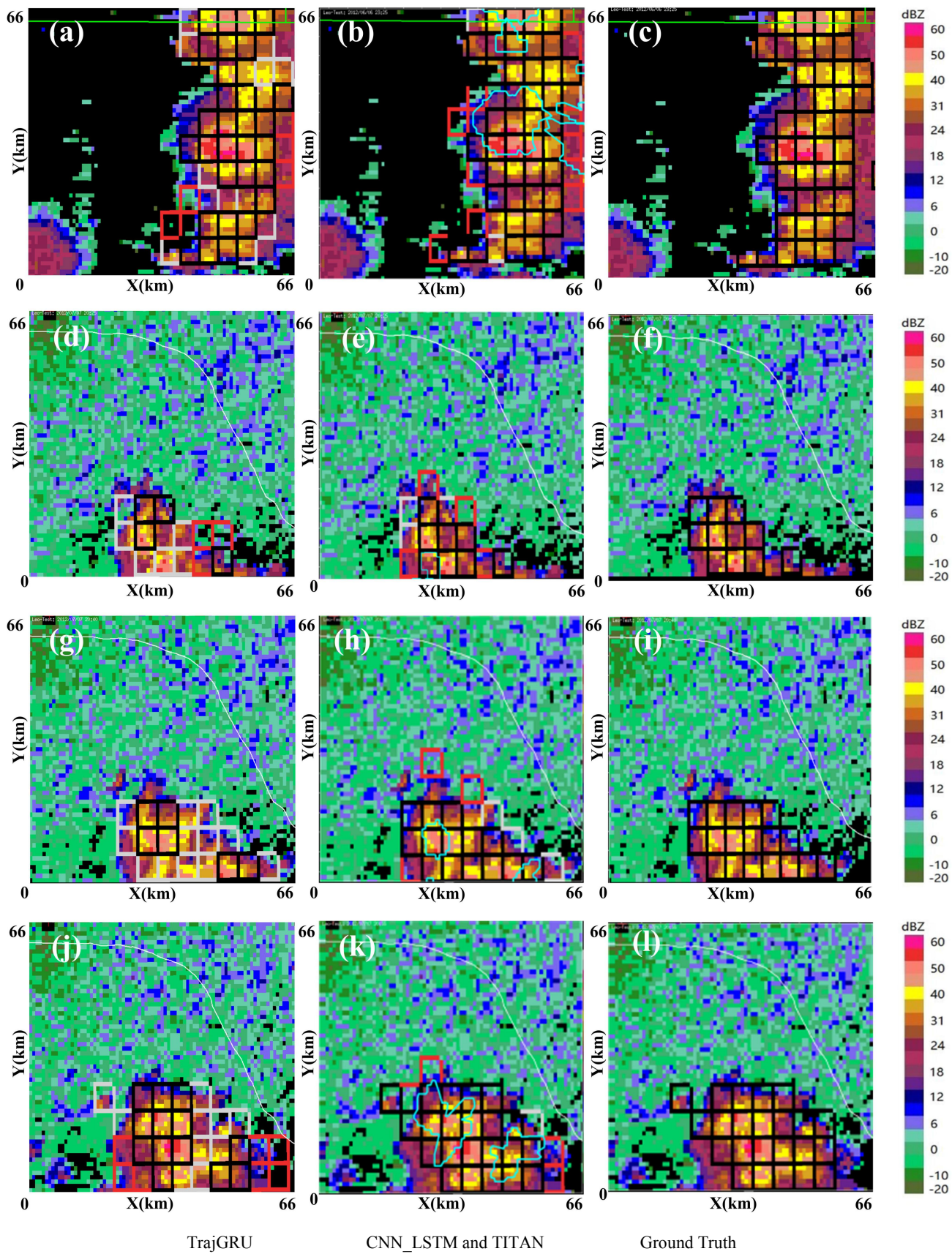

Fig. 9. Four examples for the convective storm nowcasting problem. The nowcasting time of each row is (a-c) 23:55 UTC on 6 July 2012, (d-f) 20:55 UTC on 7 July 2012, (g-i) 21:10 UTC on 7 July 2012 and (j-l) 21:25 UTC on 7 July 2012.

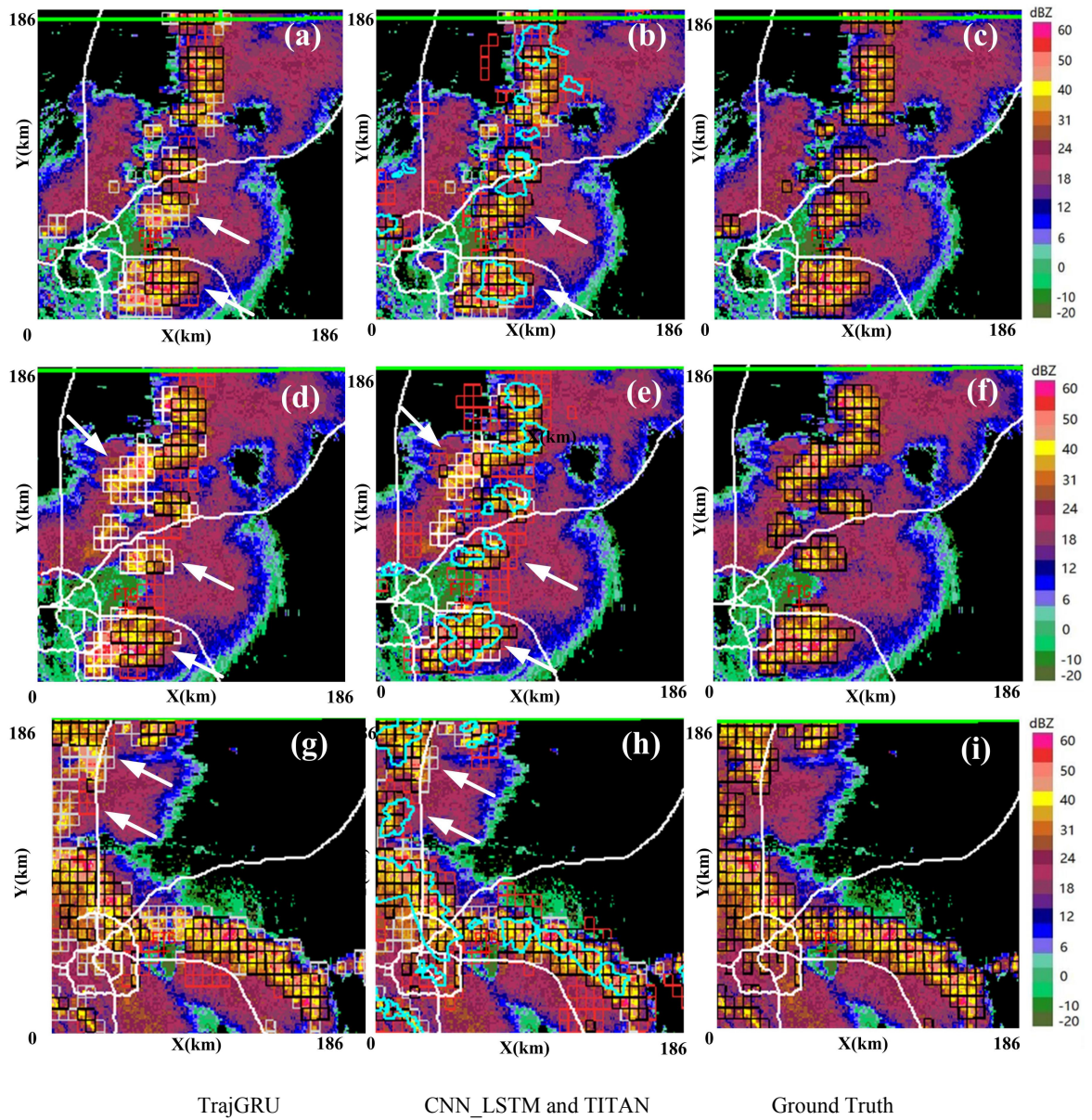

TrajGRU  CNN_LSTM and TITAN  Ground Truth

Fig. 10. Similar as Fig. 9 but three examples of nowcasting results for line storms. The nowcasting time of each row is (a-c) 00:10 UTC on 7 June 2012, (d-f) 00:25 UTC on 7 June 2012, and (g-i) 23:55 UTC on 7 July 2012.

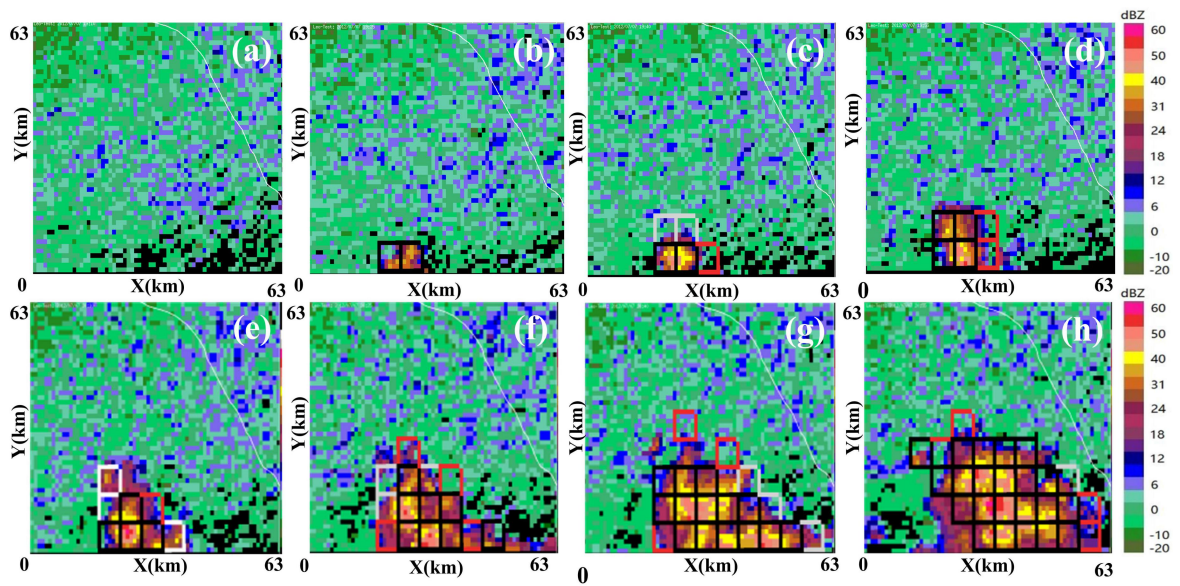

Fig. 11. The CNN_LSTM forecast results for the convective storm initiation and rapid growing process. The nowcasting time is (a) 19:40 UTC on 7 July 2012, (b) 19:55 UTC on 7 July 2012, (c) 20:10 UTC on 7 July 2012, (d) 20:25 UTC on 7 July 2012, (e) 20:40 UTC on 7 July 2012, (f) 20:55 UTC on 7 July 2012, (g) 21:10 UTC on 7 July 2012, and (h) 21:25 UTC on 7 July 2012, respectively.